\def\year{2019}\relax
\documentclass[letterpaper]{article} 
\usepackage{aaai19}  
\usepackage{times}  
\usepackage{helvet}  
\usepackage{courier}  
\usepackage{url}  
\usepackage{graphicx}  
\usepackage{epstopdf}
\usepackage{epsfig}
\usepackage{amsmath}
\DeclareMathAlphabet{\pazocal}{OMS}{zplm}{m}{n}
\newcommand{\Lb}{\pazocal{L}}
\usepackage{enumitem}
   
\usepackage{graphicx}
\usepackage{tabularx}
\usepackage{adjustbox}
\usepackage{amssymb}
\usepackage{multirow}
\usepackage[flushleft]{threeparttable}
\usepackage{subfigure}
\usepackage{color}

\newcommand{\figref}[1]{Fig.~\ref{#1}}
\newcommand{\tabref}[1]{Table~\ref{#1}}
\newcommand{\eqnref}[1]{Eq.~(\ref{#1})}

\newcommand{\etal}{\textit{et al.}}

\begin{document}
%
\title{Discriminative Feature Learning for Unsupervised Video Summarization}
\author{Yunjae Jung, Donghyeon Cho, Dahun Kim, Sanghyun Woo, In So Kweon\\
Korea Advanced Institute of Science and Technology,  Korea\\
\{yun9298a, cdh12242\}@gmail.com, \{mcahny, shwoo93, iskweon77\}@kaist.ac.kr\\
}
\maketitle
\begin{abstract}
In this paper, we address the problem of unsupervised video summarization that automatically extracts key-shots from an input video. Specifically, we tackle two critical issues based on our empirical observations: (i) Ineffective feature learning due to flat distributions of output importance scores for each frame, and (ii) training difficulty when dealing with long-length video inputs. To alleviate the first problem, we propose a simple yet effective regularization loss term called variance loss. The proposed variance loss allows a network to predict output scores for each frame with high discrepancy which enables effective feature learning and significantly improves model performance. For the second problem, we design a novel two-stream network named Chunk and Stride Network (CSNet) that utilizes local (chunk) and global (stride) temporal view on the video features. Our CSNet gives better summarization results for long-length videos compared to the existing methods. In addition, we introduce an attention mechanism to handle the dynamic information in videos. We demonstrate the effectiveness of the proposed methods by conducting extensive ablation studies and show that our final model achieves new state-of-the-art results on two benchmark datasets.

\end{abstract}

\section{Introduction}


Video has become a highly significant form of visual data, and the amount of video content uploaded to various online platforms has increased dramatically in recent years. In this regard, efficient ways of handling video have become increasingly important. One popular solution is to summarize videos into shorter ones without missing semantically important frames. Over the past few decades, many studies~\cite{song2015tvsum,ngo2003automatic,lu2013story,kim2014reconstructing,khosla2013large} have attempted to solve this problem. Recently, Zhang~\etal~showed promising results using deep neural networks, and a lot of follow-up work has been conducted in areas of supervised~\cite{zhang2016summary,zhang2016video,zhao2017hierarchical,zhao2018hsa,wei2018video} and unsupervised learning~\cite{Mahasseni2017VAEGAN,zhou2017deep}.

Supervised learning methods~\cite{zhang2016summary,zhang2016video,zhao2017hierarchical,zhao2018hsa,wei2018video} utilize ground truth labels that represent importance scores of each frame to train deep neural networks. Since human-annotated data is used, semantic features are faithfully learned. However, labeling for many video frames is expensive, and overfitting problems frequently occur when there is insufficient label data. These limitations can be mitigated by using the unsupervised learning method as in~\cite{Mahasseni2017VAEGAN,zhou2017deep}. However, since there is no human labeling in this method, a method for supervising the network needs to be appropriately designed.

Our baseline method~\cite{Mahasseni2017VAEGAN} uses a variational autoencoder (VAE)~\cite{kingma2013auto} and generative adversarial networks (GANs)~\cite{goodfellow2014generative} to learn video summarization without human labels. The key idea is that a good summary should reconstruct original video seamlessly. Features of each input frame obtained by convolutional neural network (CNN) are multiplied with predicted importance scores. Then, these features are passed to a generator to restore the original features. The discriminator is trained to distinguish between the generated (restored) features and the original ones.

Although it is fair to say that a good summary can represent and restore original video well, original features can also be restored well with uniformly distributed frame level importance scores. This trivial solution leads to difficulties in learning discriminative features to find key-shots. Our approach works to overcome this problem. When output scores become more flattened, the variance of the scores tremendously decreases. From this mathematically obvious fact, we propose a simple yet powerful way to increase the variance of the scores. Variance loss is simply defined as a reciprocal of variance of the predicted scores. 

In addition, to learn more discriminative features, we propose Chunk and Stride Network (CSNet) that simultaneously utilizes local (chunk) and global (stride) temporal views on the video. CSNet splits input features of a video into two streams (chunk and stride), then passes both split features to bidirectional long short-term memory (LSTM) and merges them back to estimate the final scores. Using chunk and stride, the difficulty of feature learning for long-length videos is overcome.

\begin{figure*}[t]
\centering
\includegraphics[width=0.85\textwidth]{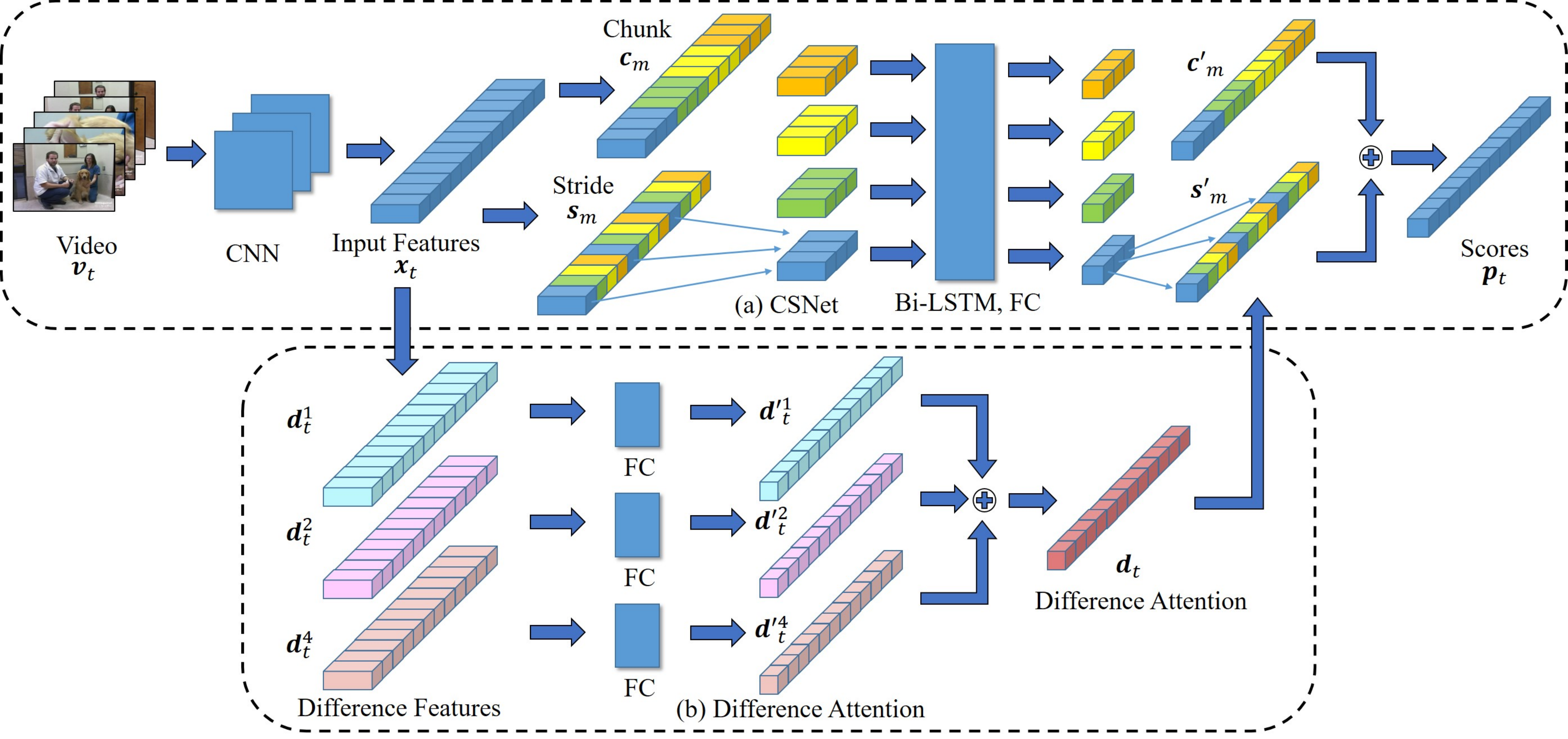}
\caption{The overall architecture of our network. (a) chunk and stride network (CSNet) splits input features $x_t$ into $c_t$ and $s_t$ by chunk and stride methods. Each orange, yellow, green, and blue color represents how the chunk and stride divide the input features $x_t$. Divided features are combined in the original order after going through LSTM and FC separately. (b) Difference attention is a approach for designing dynamic scene transitions at different temporal strides. $d^1_t$, $d^2_t$, $d^4_t$ are difference of input features $x_t$ with 1, 2, 4 temporal strides. Each difference features are summed after FC, which is denoted as difference attention $d_t$, and summed again with $c'_t$ and $s'_t$, respectively.}
\label{fig:overview}
\end{figure*}

Finally, we develop an attention mechanism to capture dynamic scene transitions, which are highly related to key-shots. In order to implement this module, we use temporal difference between frame-level CNN features. If a scene changes only slightly, the CNN features of the adjacent frames will have similar values. In contrast, at scene transitions in videos, CNN features in the adjacent frames will differ a lot. The attention module is used in conjunction with CSNet as shown in~\figref{fig:overview}, and helps to learn discriminative features by considering information about dynamic scene transitions.




We evaluate our network by conducting extensive experiments on SumMe~\cite{gygli2014creating} and TVSum~\cite{song2015tvsum} datasets. YouTube and OVP~\cite{de2011vsumm} datasets are used for the training process in augmented and transfer settings. We also conducted an ablation study to analyze the contribution of each component of our design. 
Quantitative results show the selected key-shots and demonstrate the validity of difference attention. Similar to previous methods, we randomly split the test set and the train set five times. To make the comparison fair, we exclude duplicated or skipped videos in the test set.

Our overall contributions are as follows. (i) We propose variance loss, which effectively solves the flat output problem experienced by some of the previous methods. This approach significantly improves performance, especially in unsupervised learning. (ii) We construct CSNet architecture to detect highlights in local (chunk) and global (stride) temporal view on the video. We also impose a difference attention approach to capture dynamic scene transitions which are highly related to key-shots. (iii) We analyze our methods with ablation studies and achieve the state-of-the-art performances on SumMe and TVSum datasets.

\section{Related Work}

Given an input video, video summarization aims to produce a shortened version
that highlights the representative video frames. Various
prior work has proposed solutions to this problem, including video time-lapse~\cite{joshi2015real,kopf2014first,poleg2015egosampling}, synopsis~\cite{pritch2008nonchronological}, montage~\cite{kang2006space,sun2014salient} and storyboards~\cite{gong2014diverse,gygli2014creating,gygli2015video,lee2012discovering,liu2010hierarchical,yang2015unsupervised,gong2014diverse}. Our work is most closely related to storyboards, selecting some important pieces of information to summarize key events present in the entire video.

Early work on video summarization problems heavily relied on hand-crafted features and unsupervised learning. Such work defined various heuristics to represent the importance of the frames~\cite{song2015tvsum,ngo2003automatic,lu2013story,kim2014reconstructing,khosla2013large} and to use the scores to select representative frames to build the summary video. Recent work has explored supervised learning approach for
this problem, using training data consisting
of videos and their ground-truth summaries generated by humans. These
supervised learning methods outperform early work on unsupervised
approach, since they can better learn the high-level semantic knowledge that is
used by humans to generate summaries.

Recently, deep learning based methods~\cite{zhang2016video,Mahasseni2017VAEGAN,sharghi2017query} have gained attention for video summarization tasks. The most recent studies adopt recurrent models such as LSTMs, based on the intuition that using LSTM enables the capture of long-range temporal dependencies among video frames which are critical for effective summary generation. 

Zhang~\etal~\cite{zhang2016video} introduced two LSTMs to model the variable range dependency in video summarization. One LSTM was used for video frame sequences in the forward direction, while the other LSTM was used for the backward direction. In addition, a determinantal point process model~\cite{gong2014diverse,zhang2016summary} was adopted for further improvement of diversity in the subset selection. Mahasseni~\etal.~\cite{Mahasseni2017VAEGAN} proposed an unsupervised method that was based on a generative adversarial framework. The model consists of the summarizer and discriminator. The summarizer was a variational autoencoder LSTM, which first summarized video and then reconstructed the output. The discriminator was another LSTM that learned to distinguish between its reconstruction and the input video. 

In this work, we focus on unsupervised video summarization, and adopt LSTM following previous work. However, we empirically worked out that these LSTM-based models have inherent limitations for unsupervised video summarization. In particular, two main issues exits: First, there is ineffective feature learning due to flat distribution of output importance scores and second, there is the training difficulty with long-length video inputs. To address these problems, we propose a simple yet effective regularization loss term called Variance Loss, and design a novel two-stream network named the Chunk and Stride Network. We experimentally verify that our final model considerably outperforms state-of-the-art unsupervised video summarization. The following section gives a detailed description of our method.














\section{Proposed Approach}
In this section, we introduce methods for unsupervised video summarization. Our methods are based on a variational autoencoder (VAE) and generative adversarial networks (GAN) as~\cite{Mahasseni2017VAEGAN}. We firstly deal with discriminative feature learning under a VAE-GAN framework by using variance loss. Then, a chunk and stride network (CSNet) is proposed to overcome the limitation of most of the existing methods, which is the difficulty of learning for long-length videos. CSNet resolves this problem by taking a local (chunk) and a global (stride) view of input features. Finally, to consider which part of the video is important, we use the difference in CNN features between adjacent or wider spaced video frames as attention, assuming that dynamic plays a large role in selecting key-shots. \figref{fig:overview} shows the overall structure of our proposed approach.


\subsection{Baseline Architecture}
We adopt ~\cite{Mahasseni2017VAEGAN} as our baseline, using a variational autoencoder (VAE) and generative adversarial networks (GANs) to perform unsupervised video summarization. The key idea is that a good summary should reconstruct original video seamlessly and adopt a GAN framework to reconstruct the original video from summarized key-shots.

In the model, an input video is firstly forwarded through the backbone CNN (i.e., GoogleNet), Bi-LSTM, and FC layers (encoder LSTM) to output the importance scores of each frame. The scores are multiplied with input features to select key-frames. Original features are then reconstructed from those frames using the decoder LSTM. Finally, a discriminator distinguishes whether it is from an original input video or from reconstructed ones. By following Mahasseni~\etal{}'s overall concept of VAE-GAN, we inherit the advantages, while developing our own ideas, significantly overcoming the existing limitations.


\subsection{Variance Loss}
The main assumption of our baseline~\cite{Mahasseni2017VAEGAN} is ``well-picked key-shots can reconstruct the original image well". However, for reconstructing the original image, it is better to keep all frames instead of selecting only a few key-shots. In other words, mode collapse occurs when the encoder LSTM attempts to keep all frames, which is a trivial solution. This results in flat importance output scores for each frame, which is undesirable. To prevent the output scores from being a flat distribution, we propose a variance loss as follows:

\begin{eqnarray}
\Lb_{V}(\textbf{\textit{p}}) = \frac{1}{\hat{V}(\textbf{\textit{p}}) + \textit{eps}},
\label{equ:var_loss}
\end{eqnarray}
where $\textbf{\textit{p}}= \left \{ p_{t} : t = 1,..., T \right \}$, \textit{eps} is epsilon, and $\hat{V}$$(\cdot)$ is the variance operator. $p_t$ is an output importance score at time $t$, and $T$ is the number of frames. By enforcing~\eqnref{equ:var_loss}, the network makes the difference in output scores per frames larger, then avoids a trivial solution (flat distribution).

In addition, in order to deal with outliers, we extend variance loss in~\eqnref{equ:var_loss} by utilizing the median value of scores. The variance is computed as follows:
\begin{eqnarray}
\hat{V}_{median}((\textbf{\textit{p}})) =  \frac{\sum \limits_{t=1}^{T} {|p_t - med(\textbf{\textit{p}})|^2}}{T},
\label{equ:var_loss_med}
\end{eqnarray}
where $med(\cdot)$ is the median operator. As has been reported for many years~\cite{Pratt1975medianfilter,Huang1979median,Zhang2014Wmedian}, the median value is usually more robust to outliers than the mean value. We call this modified function variance loss for the rest of the paper, and use it for all experiments.

\subsection{Chunk and Stride Network}
To handle long-length videos, which are difficult for LSTM-based methods, our approach suggests a chunk and stride network (CSNet) as a way of jointly considering a local and a global view of input features. For each frame of the input video $\textbf{\textit{v}}= \left \{ v_{t} : t = 1,..., T \right \}$, we obtain the deep features $\textbf{\textit{x}}= \left \{ x_{t} : t = 1,..., T \right \}$ of the CNN which is GoogLeNet pool-5 layer.

As shown in~\figref{fig:overview} (a), CSNet takes a long video feature $\textbf{\textit{x}}$ as an input, and divides it into  smaller sequences in two ways. The first way involves dividing $\textbf{\textit{x}}$ into successive frames, and the other way involves dividing it at a uniform interval. The streams are denoted as $\textbf{\textit{$c_{m}$}}$, and $\textbf{\textit{$s_{m}$}}$, where $\left \{m = 1,..., M \right \}$ and $M$ is the number of divisions. Specifically, $\textbf{\textit{$c_{m}$}}$ and $\textbf{\textit{$s_{m}$}}$ can be explained as follows:
\begin{eqnarray}
c_{m} = \left \{ x_{i} : i = (m-1)\cdot(\frac{T}{M}) +1,..., m \cdot (\frac{T}{M}) \right \},\\
s_{m} = \left \{ x_{i} : i = m, m+k, m+2k, ...., m+T-M  \right \},
\label{equ:csnet1}
\end{eqnarray}
where $k$ is the interval such that $k=M$. Two different sequences, $c_{m}$ and $s_{m}$, pass through the chunk and stride stream separately. Each stream consists of bidirectional LSTM (Bi-LSTM) and a fully connected (FC) layer, which predicts importance scores at the end. Then, each of the outputs are reshaped into $\textbf{\textit{$c_{m}'$}}$ and $\textbf{\textit{$s_{m}'$}}$, enforcing the maintenance of the original frame order. Then, $\textbf{\textit{$c_{m}'$}}$ and $\textbf{\textit{$s_{m}'$}}$ are added with difference attention $d_t$. Details of the attentioning process are described in the next section. The combined features are then passed through sigmoid function to predict the final scores $p_t$ as follows:
\begin{eqnarray}
p^1_t = \textit{sigmoid}\Big(c'_{t} + d_t\Big),\\
p^2_t = \textit{sigmoid}\Big(s'_{t} + d_t\Big),\\
p_t = W[p^1_t + p^2_t].
\label{equ:csnet3}
\end{eqnarray}
where $W$ is learnable parameters for weighted sum of $p^1_t$ and $p^2_t$, which allows for flexible fusion of local (chunk) and global (stride) view of input features.


\subsection{Difference Attention}

In this section, we introduce the attention module, exploiting dynamic information as guidance for the video summarization. In practice, we use the differences in CNN features of adjacent frames. The feature difference softly encodes temporally different dynamic information which can be used as a signal for deciding whether a certain frame is relatively meaningful or not.

As shown in~\figref{fig:overview} (b), the differences $d^1_t$, $d^2_t$, $d^4_t$ between $x_{t+k}$, and $x_t$ pass through the FC layer ($d'^1_t$, $d'^2_t$, $d'^4_t$) and are merged to become $d_t$, then added to both $c_{m}$ and $s_{m}$. The proposed attention modules are represented as follows:
\begin{eqnarray}
d_{1t} = |x_{t+1} - x_t|,\\
d_{2t} = |x_{t+2} - x_t|,\\
d_{4t} = |x_{t+4} - x_t|,\\
d_t = d'_{1t} + d'_{2t} + d'_{4t}.
\label{equ:diff}
\end{eqnarray}

While the difference between the features of adjacent frames can model the simplest dynamic, the wider temporal stride can include a relatively global dynamic between the scenes.


\begin{figure*}[t]
\begin{center}
\def\arraystretch{1.1}
\begin{tabular}{@{}c@{\hskip 0.01\linewidth}c@{}}

    \includegraphics[width=0.48\linewidth]{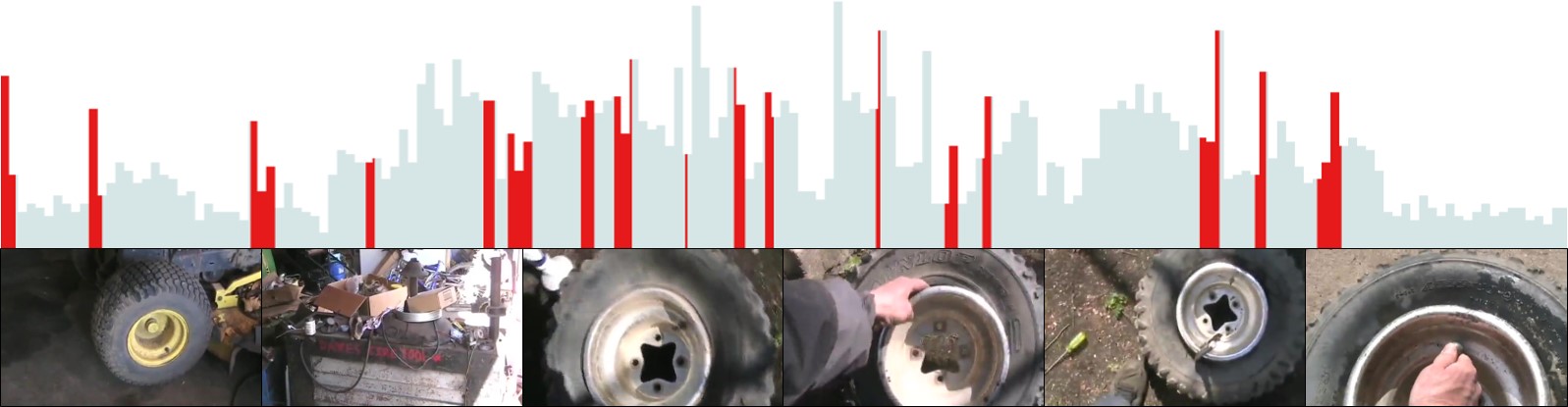}&
    \includegraphics[width=0.48\linewidth]{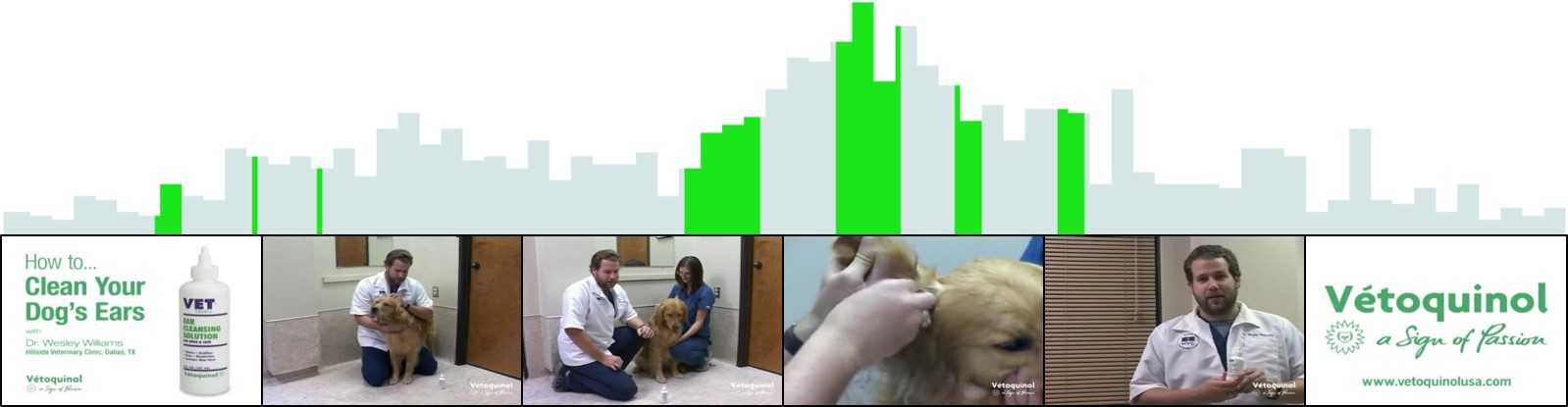}\\

{(a) Video 1} & {(b) Video 15}\\

    \includegraphics[width=0.48\linewidth]{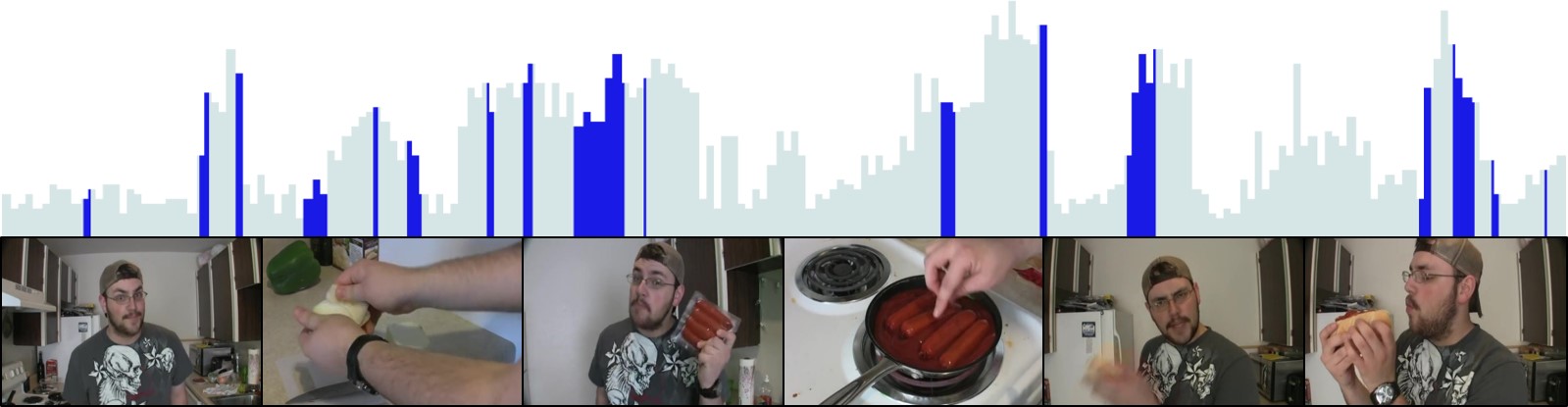}&
    \includegraphics[width=0.48\linewidth]{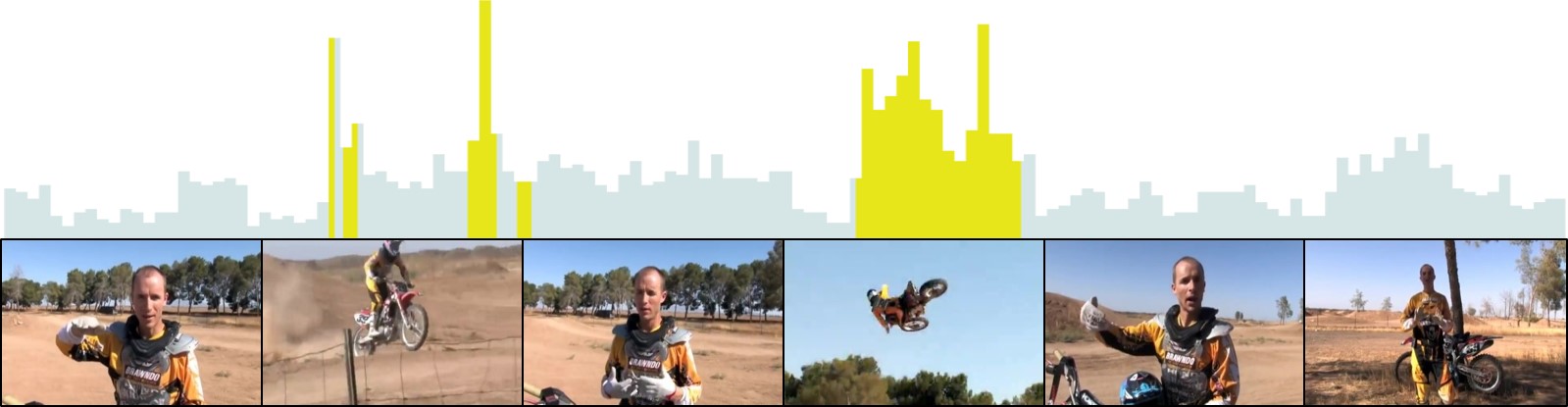}\\

{(c) Video 18} & {(d) Video 41}
\end{tabular}
\end{center}
\caption{Visualization of which key-shots are selected in the various videos of TVSum dataset. The light blue bars represent the labeled scores. Our key-shots are painted in red, green, blue, and yellow respectively in (a) - (d).}
\label{fig:visual}
\end{figure*}

\section{Experiments}

\subsection{Datasets}
We evaluate our approach on two benchmark datasets, SumMe~\cite{gygli2014creating} and TVSum~\cite{song2015tvsum}. SumMe contains 25 user videos with various events. The videos include both cases where the scene changes quickly or slowly. The length of the videos range from 1 minute to 6.5 minutes. Each video has an annotation of mostly 15 user annotations, with a maximum of 18 users. TVSum contains 50 videos with lengths ranging from 1.5 to 11 minutes. Each video in TVSum is annotated by 20 users. The annotations of SumMe and TVSum are frame-level importance scores, and we follow the evaluation method of ~\cite{zhang2016video}. OVP~\cite{de2011vsumm} and YouTube~\cite{de2011vsumm} datasets consist of 50 and 39 videos, respectively. We use OVP and YouTube datasets for transfer and augmented settings.

\subsection{Evaluation Metric}
Similar to other methods, we use the F-score used in ~\cite{zhang2016video} as an evaluation metric. In all datasets, user annotation and prediction are changed from frame-level scores to key-shots using the KTS method in ~\cite{zhang2016video}. The precision, recall, and F-score are calculated as a measure of how much the key-shots overlap. Let ``predicted" be the length of the predicted key-shots, ``user annotated" be the length of the user annotated key-shots and ``overlap" be the length of the overlapping key-shots in the following equations.
\begin{eqnarray}
P=\frac{\text{overlap}}{\text{predicted}},
R=\frac{\text{overlap}}{\text{user annotated}},\\
\text{F-score}=\frac{2PR}{P+R} * 100\%.
\label{equ:metric}
\end{eqnarray}

\begin{table}
\centering
\resizebox{1.0\linewidth}{!}{%
\begin{tabular}{ c | c | c }
\hline
Setting & Training	set & Test set \\
\hline
Canonical &	80\% SumMe &	20\% SumMe\\
Augmented & OVP + YouTube + TVSum + 80\% SumMe &	20\% SumMe\\
Transfer & OVP + YouTube + TVSum & SumMe\\
\hline
\end{tabular}
}
\caption{Evaluation setting for SumMe. In the case of TVSum, we switch between SumMe and TVSum in the above table.}
\label{tab:setting}
\end{table}

\subsection{Evaluation Settings}
Our approach is evaluated using the Canonical (C), Augmented (A), and Transfer (T) settings shown in~\tabref{tab:setting} in~\cite{zhang2016video}. To divide the test set and the training set, we randomly extract the test set five times, 20\% of the total. The remaining 80\% of the videos is used for the training set. We use the final F-score, which is the average of the F-scores of the five tests. However, if a test set is randomly selected, there may be video that is not used in the test set or is used multiple times in duplicate, making it difficult to evaluate fairly. To avoid this problem, we evaluate all the videos in the datasets without duplication or exception.

\subsection{Implementation Details}
For input features, we extract each frame by 2fps as in ~\cite{zhang2016video}, and then obtain a feature with 1024 dimensions through GoogLeNet pool-5~\cite{szegedy2015going} trained on ImageNet~\cite{russakovsky2015imagenet}. The LSTM input and hidden size is 256 reduced by FC (1024 to 256) for fast convergence, and the weight is shared with each chunk and stride input. The maximum epoch is 20, the learning rate is 1e-4, and 0.1 times after 10 epochs. The weights of the network are randomly initialized. M in CSNet is experimentally picked as 4. We implement our method using Pytorch.

\paragraph{Baseline}
Our baseline~\cite{Mahasseni2017VAEGAN} uses the VAE and GAN in the model of Mahasseni~\etal~We use their adversarial framework, which allows us unsupervised learning. Specifically, basic sparsity loss, reconstruction loss, and GAN loss are adopted. For supervised learning, we add binary cross entropy (BCE) loss between ground truth scores and predicted scores. We also put fake input, which has uniform distribution.

\subsection{Quantitative Results}
In this section, we show the experimental results of our various approach proposed in the ablation study. Then, we compare our methods with the existing unsupervised and supervised methods and finally show the experimental results in canonical, augmented, and transfer settings. For fair comparison, we quote performances of previous research recorded in~\cite{zhou2017deep}.

\begin{table}
\centering
\resizebox{1.0\linewidth}{!}{%
\begin{tabular}{ c | c | c | c | c}
\hline
Exp. & CSNet	  	& Difference & Variance Loss &	F-score (\%) \\
\hline
1 &				&				&				& 40.8\\
\hline
2 &\checkmark	&				&				& 42.0\\
3 &				&\checkmark		&				& 42.0\\
4 &				&				&\checkmark		& 44.9\\
\hline
5 &\checkmark	&\checkmark		&				& 43.5\\
6 &\checkmark	&				&\checkmark		& 49.1\\
7 &				&\checkmark		&\checkmark		& 46.9\\
\hline\hline
8 &\checkmark	&\checkmark		&\checkmark		& \textbf{51.3}\\
\hline
\end{tabular}
}
\caption{F-score (\%) of all cases where each proposed methods can be applied. When CSNet is not applied, LSTM without chunk and stride is used. Variance loss and difference attention can be simply on/off. This experiment uses SumMe dataset, unsupervised learning and canonical setting.}
\label{tab:ablation}
\end{table}

\begin{figure*}[t]
\begin{center}
\def\arraystretch{0.9}
\begin{tabular}{@{}c@{\hskip 0.01\linewidth}c@{}}

    \includegraphics[width=0.48\linewidth]{./q11_v2.jpg}&
    \includegraphics[width=0.48\linewidth]{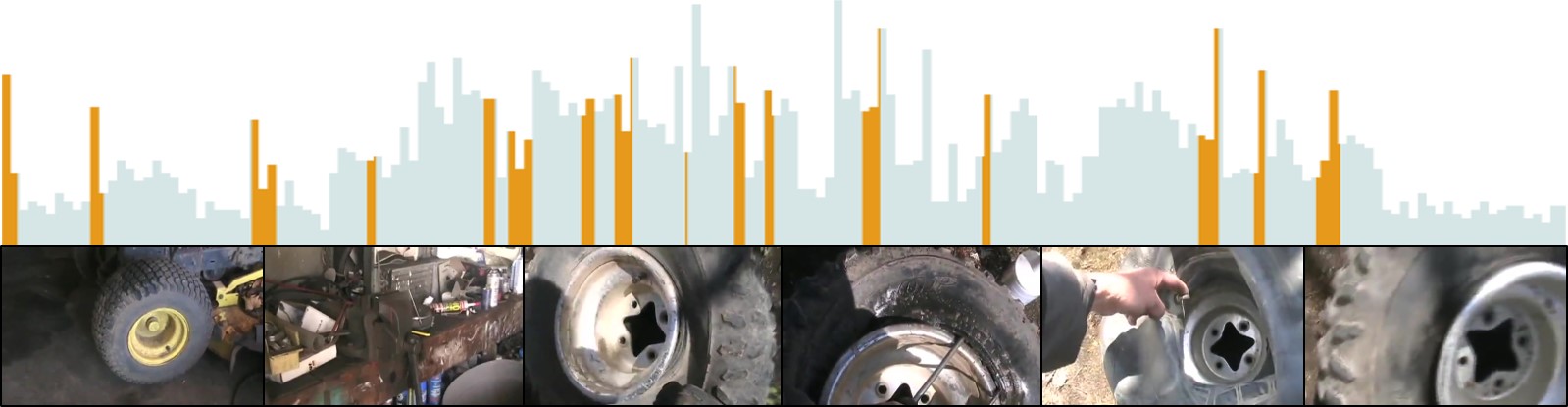}\\

{(a) CSNet 8} & {(b) CSNet 2}\\

    \includegraphics[width=0.48\linewidth]{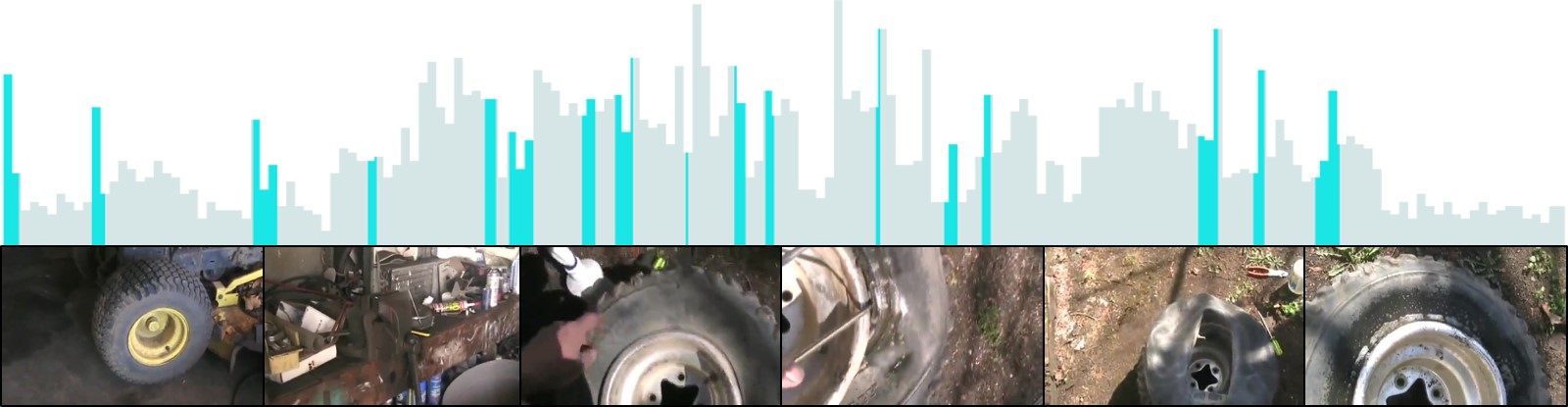}&
    \includegraphics[width=0.48\linewidth]{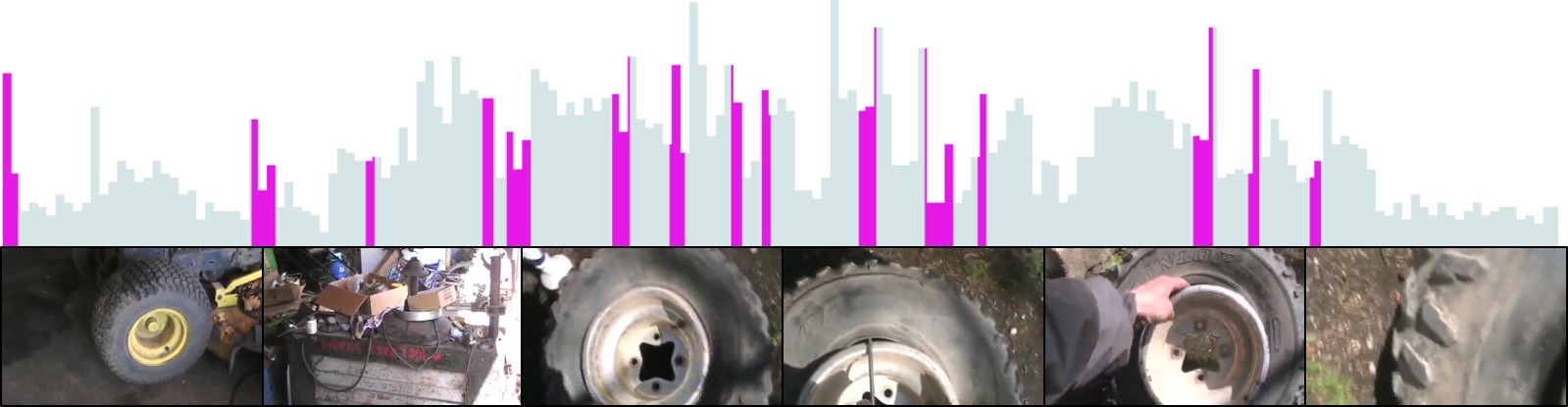}\\

{(c) CSNet 3} & {(d) CSNet 4}
\end{tabular}
\end{center}
\caption{Similar to \figref{fig:visual}, key-shots are selected by variants of CSNet denoted in ablation study. A video 1 in TVSum is used.}
\label{fig:visual-2}
\end{figure*}

\paragraph{Ablation study.}
We have three proposed approaches: CSNet, difference attention and variance loss. When all three methods are applied, the highest performance can be obtained. The ablation study in \tabref{tab:ablation} shows the contribution of each proposed method to the performance by conducting experiments on the number of cases in which each method can be applied. We call these methods shown in exp. 1 to exp. 8 CSNet\textsubscript{1} through CSNet\textsubscript{8}, respectively. If any of our proposed methods is not applied, we experiment with a version of the baseline in that we reproduce and modify some layers and hyper parameters. In this case, the lowest F-score is shown, and it is obvious that performance increases gradually when each method is applied. 

Analyzing the contribution to each method, first of all, the performance improvement due to variance loss is immensely large, which proves that it is a way to solve the problem of our baseline precisely. CSNet\textsubscript{4} is higher than CSNet\textsubscript{1} by 4.1\%, and CSNet\textsubscript{8} is better than CSNet\textsubscript{5} by 7.8\%. The variance of output scores is less than 0.001 without variance loss, but as it is applied, the variance increases to around 0.1. Since we use a reciprocal of variance to increase variance, we can observe the loss of an extremely large value in the early stages of learning. Immediately after, the effect of the loss increases the variance as a faster rate, giving the output a much wider variety of values than before.

By comparing the performance with and without the difference attention, we can see that difference attention is well modeled in the relationship between static or dynamic scene changes and frame-level importance scores. By comparing CSNet\textsubscript{1} to CSNet\textsubscript{3}, the F-score is increased by 1.2\%. Similarly, CSNet\textsubscript{5} and CSNet\textsubscript{7} are higher than CSNet\textsubscript{2} and CSNet\textsubscript{4} by 1.5\% and 2.0\%. CSNet\textsubscript{8} is greater than CSNet\textsubscript{6} by 2.2\%. These comparisons mean that the difference attention always contributes to these four cases.

We can see from our \tabref{tab:ablation} that CSNet also contributes to performance, and it is effective to design the concept of local and global features with chunk and stride while reducing input size of LSTM in temporal domain. Experiments on the number of cases where CSNet can be removed are as follow. CSNet\textsubscript{2} is better than CSNet\textsubscript{1} by 1.2\%, and each CSNet\textsubscript{5}, CSNet\textsubscript{6} outperform CSNet\textsubscript{3}, CSNet\textsubscript{4} by 1.5\%, 4.2\%. Lastly, CSNet\textsubscript{8} and CSNet\textsubscript{7} have 4.4\% difference.

Since each method improves performance as it is added, the three proposed approaches contribute individually to performance. With the combination of the proposed methods, CSNet\textsubscript{8} achieves a higher performance improvement than the sum of each F-score increased by CSNet\textsubscript{2}, CSNet\textsubscript{3} and CSNet\textsubscript{4}. In the rest of this section, we use CSNet\textsubscript{8}.

\begin{table}
\centering
\resizebox{0.8\linewidth}{!}{%
\begin{tabular}{ l | c | c }
\hline
Method & SumMe	& TVSum \\
\hline
K-medoids &	33.4 & 28.8 \\
Vsumm & 33.7 & - \\
Web image &	- & 36.0 \\
Dictionary selection & 37.8 & 42.0  \\
Online sparse coding & - &	46.0\\
Co-archetypal & - & 50.0\\
GAN\textsubscript{dpp} & 39.1 & 51.7 \\
DR-DSN & 41.4 & 57.6 \\
\hline
\hline
CSNet & \textbf{51.3}	&   \textbf{58.8} \\ 
\hline
\end{tabular}
}
\caption{F-score (\%) of unsupervised methods in canonical setting on SumMe and TVSum datasets. Our approach outperforms other existing methods. Dramatic performance improvement is shown on the SumMe dataset.}
\label{tab:unsupervised}
\end{table}

\paragraph{Comparison with unsupervised approaches.}
\tabref{tab:unsupervised} shows the experimental results for SumMe and TVSum datasets using unsupervised learning in canonical settings. Since our approach mainly target unsupervised learning, CSNet outperforms both SumMe and TVSum over the existing methods~\cite{elhamifar2012see,khosla2013large,de2011vsumm,zhao2014quasi,song2015tvsum,zhou2017deep,Mahasseni2017VAEGAN}. As a significant improvement in performance for the SumMe dataset, \tabref{tab:unsupervised} shows a F-score enhancement over 9.9\% compared to the best of the existing methods~\cite{zhou2017deep}. 

To the best of our knowledge, all existing methods are scored at less than 50\% of the F-score in the SumMe dataset. Evaluation of the SumMe dataset is more challenging than the TVSum dataset in terms of performance. DR-DSN has already made a lot of progress for the TVSum dataset, but for the first time, we have achieved extreme advancement in the SumMe dataset which decreases the gap between SumMe and TVSum. 

An interesting observation of supervised learning in video summarization is the non-optimal ground truth scores. Users who evaluated video for each data set are different, and every user does not make a consistent evaluation. In such cases, there may be a better summary than the ground truth which is a mean value of multiple user annotations. Surprisingly, during our experiments we observe that predictions for some videos receive better F-scores than in the results of ground truth. Unsupervised approaches do not use the ground truth, so it provides a step closer to the user annotation.

\begin{table}
\centering
\resizebox{0.8\linewidth}{!}{%
\begin{tabular}{ l | c | c }
\hline
Method & SumMe	& TVSum \\
\hline
Interestingness & 39.4 & - \\
Submodularity & 39.7 &	- \\
Summary transfer &	40.9 & -  \\
Bi-LSTM & 37.6 & 54.2  \\
DPP-LSTM & 38.6	& 54.7\\
GAN\textsubscript{sup} & 41.7 &  56.3 \\
DR-DSN\textsubscript{sup} &	42.1 & 58.1\\
\hline
\hline
CSNet\textsubscript{sup} & 	\textbf{48.6} &   \textbf{58.5} \\
\hline
\end{tabular}
}
\caption{F-score (\%) of supervised methods in canonical setting on SumMe and TVSum datasets. We achieve the state-of-the-art performance.}
\label{tab:supervised}
\end{table}

\paragraph{Comparison with supervised approaches.}
We implemented CSNet\textsubscript{sup} for supervised learning by simply adding binary cross entropy loss between prediction and ground truth to existing loss for CSNet. In \tabref{tab:supervised}, CSNet\textsubscript{sup} obtains state-of-the-art results compared to existing methods~\cite{gygli2014creating,gygli2015video,zhang2016summary,zhang2016video,zhou2017deep}, but does not provide a better performance than CSNet. In general, supervision improves performance, but in our case, the point of view mentioned in the unsupervised approaches may fall out of step with using ground truth directly.

\begin{table}
\centering
\resizebox{1.0\linewidth}{!}{%
\begin{tabular}{ l | c c c | c c c }
\hline
 & & SumMe	& & & TVSum \\
\hline
Method &C & A &	T &		C &	A &	T \\
\hline
Bi-LSTM &		37.6 &	41.6 &	40.7 &		54.2 &	57.9 &	56.9 \\
DPP-LSTM &		38.6 &	42.9 &	41.8 &		54.7 &	59.6 &	58.7 \\
GAN\textsubscript{dpp} &		39.1 &	43.4 &	- &		51.7 &	59.5 &	- \\
GAN\textsubscript{sup} &		41.7 &	43.6 &	- &		56.3 &	\textbf{61.2} &	- \\
DR-DSN &		41.4 &	42.8 &	42.4 &		57.6 &	58.4 &	57.8 \\
DR-DSN\textsubscript{sup} &		42.1 &	43.9 &	42.6 &		58.1 &	59.8 &	58.9 \\
HSA-RNN &		- &	44.1 &	- &		- &	59.8 &	- \\
\hline
\hline
CSNet	  				 &	\textbf{51.3} &	\textbf{52.1}	&	\textbf{45.1} &		\textbf{58.8} &	59.0 &	\textbf{59.2} \\
CSNet\textsubscript{sup} &	48.6 &	48.7	&	44.1 &		58.5 &	57.1 &	57.4 \\
\hline
\end{tabular}
}
\caption{F-score (\%) of both unsupervised and supervised methods in canonical, augmented and transfer settings on SumMe and TVSum datasets.}
\label{tab:CAT}
\end{table}

\paragraph{Comparison in augmented and transfer settings.}
We compare our CSNet with other state-of-the-art literature with augmented and transfer settings in \tabref{tab:CAT}. We can make a fair comparison using the 256 hidden layer size of LSTM used by DR-DSN~\cite{zhou2017deep}, which is a previous state-of-the-art method. We obtain better performance in CSNet than CSNet\textsubscript{sup}, and our unsupervised CSNet performs better than the supervised method in any other approach except for GAN\textsubscript{sup}, which uses 1024 hidden size in TVSum dataset with augmented setting.

\subsection{Qualitative Results}

\paragraph{Selected key-shots.}
In this section, we visualize selected key-shots in two ways.
First, in \figref{fig:visual}, selected key-shots are visualized in bar graph form using various genre of videos. (a) - (d) show that many of our key-shots select peak points of labeled scores. In terms of the content of the video, the scenes selected by CSNet are mostly meaningful scenes by comparing colored bars with the images in \figref{fig:visual}. Then, in \figref{fig:visual-2}, we compare variants of our approach with a video 1 in TVSum. Although minor differences exist, each approach select peak points well.

\begin{figure}[t]
\centering\resizebox{0.85\linewidth}{!}{
\includegraphics[width=0.88\textwidth]{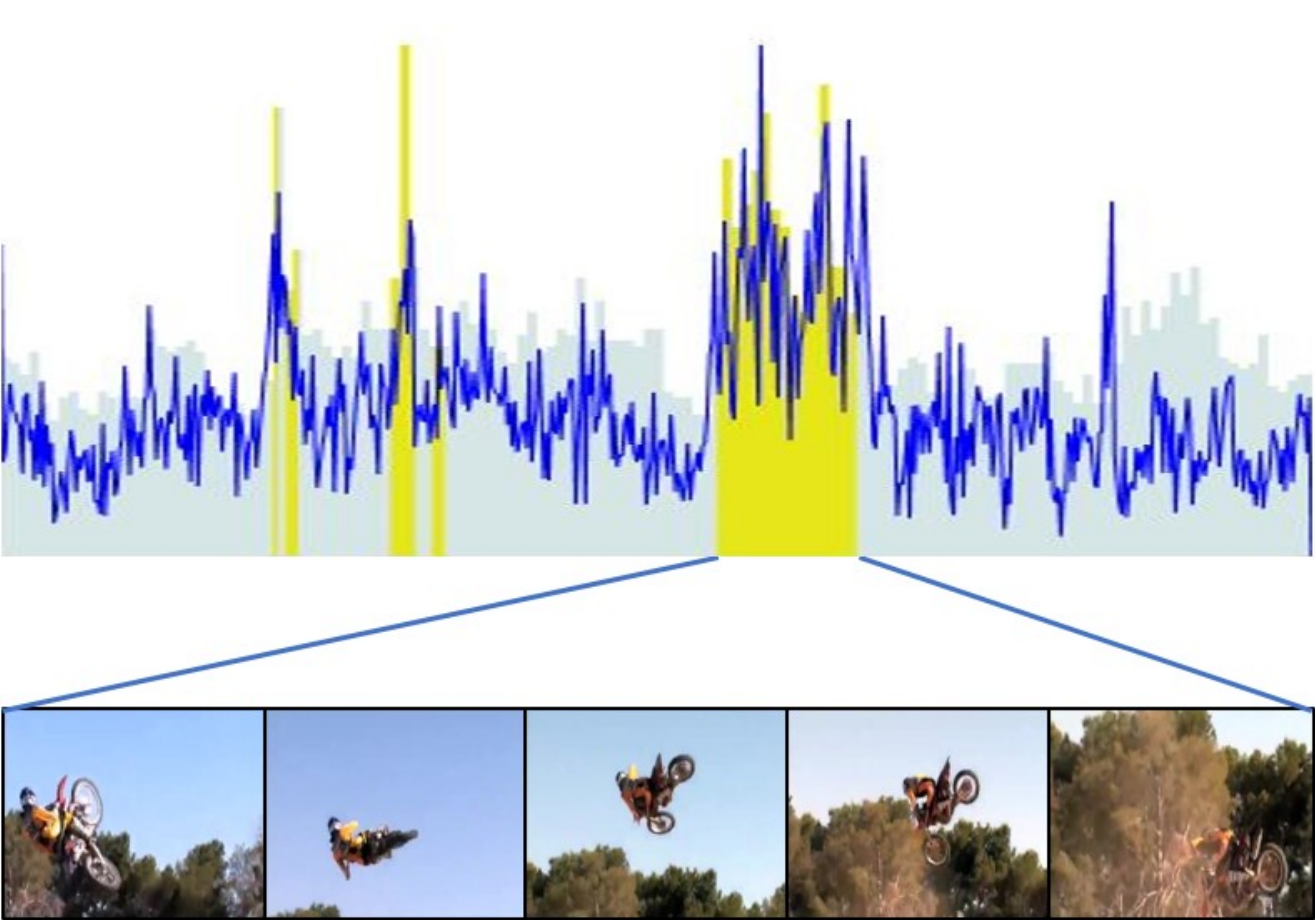}
}
\caption{ Experiment with video 41 in the TVSum dataset. In addition to the visualization results in \figref{fig:visual}, difference attention is plotted with blue color. When visualizing the difference attention, it is normalized to have a same range of ground truth scores. The picture is the video frames which are mainly predicted part with key-shots.}
\label{fig:diff}
\end{figure}

\paragraph{Difference attention.}
With a deeper analysis of difference attention, we visualize the difference attention in the TVSum dataset. Its motivation is to capture dynamic information between frames of video. We can verify our assumption that the dynamic scene should be more important than the static scene with this experiment. As shown in \figref{fig:diff}, the plotted blue graph is in line with the selected key-shots, which highlight portions with high scores. The selected key-shots are of a motorcycle jump, which is a dynamic scene in the video. As a result, difference attention can effectively predict key-shots using dynamic information.

\section{Conclusion}
In this paper, we propose discriminative feature learning for unsupervised video summarization with our approach. Variance loss tackles the temporal dependency problem, which causes a flat output problem in LSTM. CSNet designs a local and global scheme, which reduces temporal input size for LSTM. Difference attention highlights dynamic information, which is highly related to key-shots in a video. Extensive experiments on two benchmark datasets including ablation study show that our state-of-the-art unsupervised approach outperforms most of the supervised methods.

\paragraph{Acknowledgements}
This research is supported by the Study on Deep Visual Understanding funded by the Samsung Electronics Co., Ltd (Samsung Research)
\medskip


{
	\bibliographystyle{aaai} 
	\bibliography{egbib.bib}

\begin{thebibliography}{}

\bibitem[\protect\citeauthoryear{De~Avila \bgroup et al\mbox.\egroup
  }{2011}]{de2011vsumm}
De~Avila, S. E.~F.; Lopes, A. P.~B.; da~Luz~Jr, A.; and
  de~Albuquerque~Ara{\'u}jo, A.
\newblock 2011.
\newblock Vsumm: A mechanism designed to produce static video summaries and a
  novel evaluation method.
\newblock {\em Pattern Recognition Letters} 32(1):56--68.

\bibitem[\protect\citeauthoryear{Elhamifar, Sapiro, and
  Vidal}{2012}]{elhamifar2012see}
Elhamifar, E.; Sapiro, G.; and Vidal, R.
\newblock 2012.
\newblock See all by looking at a few: Sparse modeling for finding
  representative objects.
\newblock In {\em {Proc. of Computer Vision and Pattern Recognition (CVPR)}},
  1600--1607.
\newblock IEEE.

\bibitem[\protect\citeauthoryear{Gong \bgroup et al\mbox.\egroup
  }{2014}]{gong2014diverse}
Gong, B.; Chao, W.-L.; Grauman, K.; and Sha, F.
\newblock 2014.
\newblock Diverse sequential subset selection for supervised video
  summarization.
\newblock In {\em {Proc. of Neural Information Processing Systems (NIPS)}},
  2069--2077.

\bibitem[\protect\citeauthoryear{Goodfellow \bgroup et al\mbox.\egroup
  }{2014}]{goodfellow2014generative}
Goodfellow, I.; Pouget-Abadie, J.; Mirza, M.; Xu, B.; Warde-Farley, D.; Ozair,
  S.; Courville, A.; and Bengio, Y.
\newblock 2014.
\newblock Generative adversarial nets.
\newblock In {\em {Proc. of Neural Information Processing Systems (NIPS)}},
  2672--2680.

\bibitem[\protect\citeauthoryear{Gygli \bgroup et al\mbox.\egroup
  }{2014}]{gygli2014creating}
Gygli, M.; Grabner, H.; Riemenschneider, H.; and Van~Gool, L.
\newblock 2014.
\newblock Creating summaries from user videos.
\newblock In {\em {Proc. of European Conf. on Computer Vision (ECCV)}},
  505--520.
\newblock Springer.

\bibitem[\protect\citeauthoryear{Gygli, Grabner, and
  Van~Gool}{2015}]{gygli2015video}
Gygli, M.; Grabner, H.; and Van~Gool, L.
\newblock 2015.
\newblock Video summarization by learning submodular mixtures of objectives.
\newblock In {\em {Proc. of Computer Vision and Pattern Recognition (CVPR)}},
  3090--3098.

\bibitem[\protect\citeauthoryear{Huang, Yang, and Tang}{1979}]{Huang1979median}
Huang, T.; Yang, G.; and Tang, G.
\newblock 1979.
\newblock A fast two-dimensional median filtering algorithm.
\newblock {\em {IEEE Trans. on Acoustics, Speech and Signal Processing.
  (APSP)}} 27(1):13--18.

\bibitem[\protect\citeauthoryear{Joshi \bgroup et al\mbox.\egroup
  }{2015}]{joshi2015real}
Joshi, N.; Kienzle, W.; Toelle, M.; Uyttendaele, M.; and Cohen, M.~F.
\newblock 2015.
\newblock Real-time hyperlapse creation via optimal frame selection.
\newblock {\em ACM Transactions on Graphics (TOG)} 34(4):63.

\bibitem[\protect\citeauthoryear{Kang \bgroup et al\mbox.\egroup
  }{2006}]{kang2006space}
Kang, H.-W.; Matsushita, Y.; Tang, X.; and Chen, X.-Q.
\newblock 2006.
\newblock Space-time video montage.
\newblock In {\em {Proc. of Computer Vision and Pattern Recognition (CVPR)}},
  volume~2,  1331--1338.
\newblock IEEE.

\bibitem[\protect\citeauthoryear{Khosla \bgroup et al\mbox.\egroup
  }{2013}]{khosla2013large}
Khosla, A.; Hamid, R.; Lin, C.-J.; and Sundaresan, N.
\newblock 2013.
\newblock Large-scale video summarization using web-image priors.
\newblock In {\em {Proc. of Computer Vision and Pattern Recognition (CVPR)}},
  2698--2705.

\bibitem[\protect\citeauthoryear{Kim and Xing}{2014}]{kim2014reconstructing}
Kim, G., and Xing, E.~P.
\newblock 2014.
\newblock Reconstructing storyline graphs for image recommendation from web
  community photos.
\newblock In {\em {Proc. of Computer Vision and Pattern Recognition (CVPR)}},
  3882--3889.

\bibitem[\protect\citeauthoryear{Kingma and Welling}{2013}]{kingma2013auto}
Kingma, D.~P., and Welling, M.
\newblock 2013.
\newblock Auto-encoding variational bayes.
\newblock In {\em {Proc. of Int'l Conf. on Learning Representations (ICLR)}}.

\bibitem[\protect\citeauthoryear{Kopf, Cohen, and
  Szeliski}{2014}]{kopf2014first}
Kopf, J.; Cohen, M.~F.; and Szeliski, R.
\newblock 2014.
\newblock First-person hyper-lapse videos.
\newblock {\em ACM Transactions on Graphics (TOG)} 33(4):78.

\bibitem[\protect\citeauthoryear{Lee, Ghosh, and
  Grauman}{2012}]{lee2012discovering}
Lee, Y.~J.; Ghosh, J.; and Grauman, K.
\newblock 2012.
\newblock Discovering important people and objects for egocentric video
  summarization.
\newblock In {\em {Proc. of Computer Vision and Pattern Recognition (CVPR)}},
  1346--1353.
\newblock IEEE.

\bibitem[\protect\citeauthoryear{Liu, Hua, and
  Chen}{2010}]{liu2010hierarchical}
Liu, D.; Hua, G.; and Chen, T.
\newblock 2010.
\newblock A hierarchical visual model for video object summarization.
\newblock {\em {IEEE Trans. Pattern Anal. Mach. Intell. (TPAMI)}}
  32(12):2178--2190.

\bibitem[\protect\citeauthoryear{Lu and Grauman}{2013}]{lu2013story}
Lu, Z., and Grauman, K.
\newblock 2013.
\newblock Story-driven summarization for egocentric video.
\newblock In {\em {Proc. of Computer Vision and Pattern Recognition (CVPR)}},
  2714--2721.

\bibitem[\protect\citeauthoryear{Mahasseni, Lam, and
  Todorovic}{2017}]{Mahasseni2017VAEGAN}
Mahasseni, B.; Lam, M.; and Todorovic, S.
\newblock 2017.
\newblock Unsupervised video summarization with adversarial lstm networks.
\newblock In {\em {Proc. of Computer Vision and Pattern Recognition (CVPR)}},
  2982--2991.

\bibitem[\protect\citeauthoryear{Ngo, Ma, and Zhang}{2003}]{ngo2003automatic}
Ngo, C.-W.; Ma, Y.-F.; and Zhang, H.-J.
\newblock 2003.
\newblock Automatic video summarization by graph modeling.
\newblock In {\em Computer Vision, 2003. Proceedings. Ninth IEEE International
  Conference on},  104--109.
\newblock IEEE.

\bibitem[\protect\citeauthoryear{Poleg \bgroup et al\mbox.\egroup
  }{2015}]{poleg2015egosampling}
Poleg, Y.; Halperin, T.; Arora, C.; and Peleg, S.
\newblock 2015.
\newblock Egosampling: Fast-forward and stereo for egocentric videos.
\newblock In {\em {Proc. of Computer Vision and Pattern Recognition (CVPR)}},
  4768--4776.

\bibitem[\protect\citeauthoryear{Pratt}{1975}]{Pratt1975medianfilter}
Pratt, W.~K.
\newblock 1975.
\newblock Median filtering.
\newblock {\em Semiannual Report, Univ. of Southern California}.

\bibitem[\protect\citeauthoryear{Pritch, Rav-Acha, and
  Peleg}{2008}]{pritch2008nonchronological}
Pritch, Y.; Rav-Acha, A.; and Peleg, S.
\newblock 2008.
\newblock Nonchronological video synopsis and indexing.
\newblock {\em {IEEE Trans. Pattern Anal. Mach. Intell. (TPAMI)}}
  30(11):1971--1984.

\bibitem[\protect\citeauthoryear{Russakovsky \bgroup et al\mbox.\egroup
  }{2015}]{russakovsky2015imagenet}
Russakovsky, O.; Deng, J.; Su, H.; Krause, J.; Satheesh, S.; Ma, S.; Huang, Z.;
  Karpathy, A.; Khosla, A.; Bernstein, M.; et~al.
\newblock 2015.
\newblock Imagenet large scale visual recognition challenge.
\newblock {\em {Int'l Journal of Computer Vision (IJCV)}} 115(3):211--252.

\bibitem[\protect\citeauthoryear{Sharghi, Laurel, and
  Gong}{2017}]{sharghi2017query}
Sharghi, A.; Laurel, J.~S.; and Gong, B.
\newblock 2017.
\newblock Query-focused video summarization: Dataset, evaluation, and a memory
  network based approach.
\newblock In {\em {Proc. of Computer Vision and Pattern Recognition (CVPR)}},
  2127--2136.

\bibitem[\protect\citeauthoryear{Song \bgroup et al\mbox.\egroup
  }{2015}]{song2015tvsum}
Song, Y.; Vallmitjana, J.; Stent, A.; and Jaimes, A.
\newblock 2015.
\newblock Tvsum: Summarizing web videos using titles.
\newblock In {\em {Proc. of Computer Vision and Pattern Recognition (CVPR)}},
  5179--5187.

\bibitem[\protect\citeauthoryear{Sun \bgroup et al\mbox.\egroup
  }{2014}]{sun2014salient}
Sun, M.; Farhadi, A.; Taskar, B.; and Seitz, S.
\newblock 2014.
\newblock Salient montages from unconstrained videos.
\newblock In {\em {Proc. of European Conf. on Computer Vision (ECCV)}},
  472--488.
\newblock Springer.

\bibitem[\protect\citeauthoryear{Szegedy \bgroup et al\mbox.\egroup
  }{2015}]{szegedy2015going}
Szegedy, C.; Liu, W.; Jia, Y.; Sermanet, P.; Reed, S.; Anguelov, D.; Erhan, D.;
  Vanhoucke, V.; and Rabinovich, A.
\newblock 2015.
\newblock Going deeper with convolutions.
\newblock In {\em {Proc. of Computer Vision and Pattern Recognition (CVPR)}},
  1--9.

\bibitem[\protect\citeauthoryear{Wei \bgroup et al\mbox.\egroup
  }{2018}]{wei2018video}
Wei, H.; Ni, B.; Yan, Y.; Yu, H.; Yang, X.; and Yao, C.
\newblock 2018.
\newblock Video summarization via semantic attended networks.
\newblock In {\em Proc. of Association for the Advancement of Artificial
  Intelligence (AAAI)}.

\bibitem[\protect\citeauthoryear{Yang \bgroup et al\mbox.\egroup
  }{2015}]{yang2015unsupervised}
Yang, H.; Wang, B.; Lin, S.; Wipf, D.; Guo, M.; and Guo, B.
\newblock 2015.
\newblock Unsupervised extraction of video highlights via robust recurrent
  auto-encoders.
\newblock In {\em {Proc. of Int'l Conf. on Computer Vision (ICCV)}},
  4633--4641.

\bibitem[\protect\citeauthoryear{Zhang \bgroup et al\mbox.\egroup
  }{2016a}]{zhang2016summary}
Zhang, K.; Chao, W.-L.; Sha, F.; and Grauman, K.
\newblock 2016a.
\newblock Summary transfer: Exemplar-based subset selection for video
  summarization.
\newblock In {\em {Proc. of Computer Vision and Pattern Recognition (CVPR)}},
  1059--1067.

\bibitem[\protect\citeauthoryear{Zhang \bgroup et al\mbox.\egroup
  }{2016b}]{zhang2016video}
Zhang, K.; Chao, W.-L.; Sha, F.; and Grauman, K.
\newblock 2016b.
\newblock Video summarization with long short-term memory.
\newblock In {\em {Proc. of European Conf. on Computer Vision (ECCV)}},
  766--782.
\newblock Springer.

\bibitem[\protect\citeauthoryear{Zhang, Xu, and Jia}{2014}]{Zhang2014Wmedian}
Zhang, Q.; Xu, L.; and Jia, J.
\newblock 2014.
\newblock 100+ times faster weighted median filter.
\newblock In {\em {Proc. of Computer Vision and Pattern Recognition (CVPR)}},
  2830--2837.

\bibitem[\protect\citeauthoryear{Zhao and Xing}{2014}]{zhao2014quasi}
Zhao, B., and Xing, E.~P.
\newblock 2014.
\newblock Quasi real-time summarization for consumer videos.
\newblock In {\em {Proc. of Computer Vision and Pattern Recognition (CVPR)}},
  2513--2520.

\bibitem[\protect\citeauthoryear{Zhao, Li, and Lu}{2017}]{zhao2017hierarchical}
Zhao, B.; Li, X.; and Lu, X.
\newblock 2017.
\newblock Hierarchical recurrent neural network for video summarization.
\newblock In {\em {Proc. of Multimedia Conference (MM)}},  863--871.
\newblock ACM.

\bibitem[\protect\citeauthoryear{Zhao, Li, and Lu}{2018}]{zhao2018hsa}
Zhao, B.; Li, X.; and Lu, X.
\newblock 2018.
\newblock Hsa-rnn: Hierarchical structure-adaptive rnn for video summarization.
\newblock In {\em {Proc. of Computer Vision and Pattern Recognition (CVPR)}},
  7405--7414.

\bibitem[\protect\citeauthoryear{Zhou and Qiao}{2018}]{zhou2017deep}
Zhou, K., and Qiao, Y.
\newblock 2018.
\newblock Deep reinforcement learning for unsupervised video summarization with
  diversity-representativeness reward.
\newblock In {\em Proc. of Association for the Advancement of Artificial
  Intelligence (AAAI)}.

\end{thebibliography}
}

\end{document}